\documentclass[letterpaper]{article} 
\usepackage{aaai24}  
\usepackage{times}  
\usepackage{helvet}  
\usepackage{courier}  
\usepackage[hyphens]{url}  
\usepackage{graphicx} 
\usepackage{natbib}  
\usepackage{caption} 
\usepackage{algorithm}
\usepackage{algorithmic}

\usepackage{makecell}
\usepackage{booktabs}
\usepackage[table,xcdraw]{xcolor}
\usepackage{multirow}
\usepackage{amssymb}
\usepackage{amsmath}

\def\eg{e.g.} 
\def\ie{i.e.} 
\def\cf{c.f.} 

\usepackage{newfloat}
\usepackage{listings}
\DeclareCaptionStyle{ruled}{labelfont=normalfont,labelsep=colon,strut=off} 
\floatstyle{ruled}
\newfloat{listing}{tb}{lst}{}
\floatname{listing}{Listing}
\title{Weakly Supervised Semantic Segmentation for Driving Scenes}
\author{
    Dongseob Kim\equalcontrib\textsuperscript{\rm 1}, 
    Seungho Lee\equalcontrib\textsuperscript{\rm 1},
    Junsuk Choe\textsuperscript{\rm 2},
    Hyunjung Shim\thanks{Hyunjung Shim is a corresponding author.}\textsuperscript{\rm 3}
}
\affiliations{
    \textsuperscript{\rm 1} Yonsei University, South Korea\\
    \textsuperscript{\rm 2} Sogang University, South Korea\\
    \textsuperscript{\rm 3} Korea Advanced Institute of Science \& Technology, South Korea\\
    \{kou.k, seungholee\}@yonsei.ac.kr, jschoe@sogang.ac.kr, kateshim@kaist.ac.kr
}

\begin{document}

\maketitle

\begin{abstract}
State-of-the-art techniques in weakly-supervised semantic segmentation (WSSS) using image-level labels exhibit severe performance degradation on driving scene datasets such as Cityscapes. To address this challenge, we develop a new WSSS framework tailored to driving scene datasets. Based on extensive analysis of dataset characteristics, we employ Contrastive Language-Image Pre-training (CLIP) as our baseline to obtain pseudo-masks. However, CLIP introduces two key challenges: (1) pseudo-masks from CLIP lack in representing small object classes, and (2) these masks contain notable noise. We propose solutions for each issue as follows. (1) We devise Global-Local View Training that seamlessly incorporates small-scale patches during model training, thereby enhancing the model's capability to handle small-sized yet critical objects in driving scenes (\eg, \textit{traffic light}). (2) We introduce Consistency-Aware Region Balancing (CARB), a novel technique that discerns reliable and noisy regions through evaluating the consistency between CLIP masks and segmentation predictions. It prioritizes reliable pixels over noisy pixels via adaptive loss weighting. Notably, the proposed method achieves 51.8\% mIoU on the Cityscapes test dataset, showcasing its potential as a strong WSSS baseline on driving scene datasets. Experimental results on CamVid and WildDash2 demonstrate the effectiveness of our method across diverse datasets, even with small-scale datasets or visually challenging conditions. The code is available at https://github.com/k0u-id/CARB.
\end{abstract}

\section{Introduction}

Recent advancements in weakly supervised semantic segmentation (WSSS) using image-level labels have demonstrated impressive results, achieving performance levels of over 90\% compared to full supervised models on the PASCAL VOC dataset~\cite{lee2022threshold,yoon2022adversarial}. Given this success, it is crucial to transfer the WSSS framework to driving scenes, which are a significant scenario in semantic segmentation. Obtaining pixel-level labels in driving scenes is prohibitively expensive, making label-efficient training methods imperative in this context. For instance, Cityscapes required 1.5 hours per image~\cite{cordts2016cityscapes}, while PASCAL VOC required 239.7 seconds per image~\cite{bearman2016s}.

However, when applied to driving scene datasets like Cityscapes, WSSS models exhibit significant performance degradation. \citeauthor{akiva2023single} attributed this issue to the specific characteristics of the dataset, such as small object size, a high number of objects in each image, and limited diversity in object appearance~\cite{akiva2023single}. However, they only reported this tendency implicitly. In our study, we explicitly compare the driving scene datasets to the existing benchmark datasets (\ie, PASCAL VOC and MS COCO). As a result, we find that the driving scenes datasets lack negative samples and exhibit a remarkably high level of co-occurrence among classes. This poses a challenge in identifying individual objects through image classification, which hinders the effectiveness of common WSSS baselines, such as class activation mapping (CAM).

Recently, Contrastive Language Image Pre-training (CLIP), a model trained on a massive set of 400 million image-text pairs, has remarkably performed in open vocabulary classification. Using the open vocabulary classification ability of CLIP, we can avoid the characteristic of the dataset degrading the classifier's performance. As a result, as opposed to CAM, the seed mask generated by CLIP better distinguishes the object regions on the driving dataset like Cityscapes. Despite the potential, it often fails to identify small objects and produces noisy masks (Fig.~\ref{fig:mask_resizencrop}~(a)).

In this paper, we present a novel WSSS framework for driving scene datasets, to address the above two challenges inherent in CLIP. Considering CLIP as a baseline mask generator, we propose (1) global-local view training to handle small-sized objects and (2) \textit{Consistency-Aware Region Balancing (CARB)} to mitigate the negative effects of noisy pseudo-masks. Firstly, we found a unique property of CLIP: it offers considerably different pseudo-masks across input scales. Based on this observation, we use both a local view (\ie, a small-sized patch) and a global view (\ie, an original-sized image) during model training for accurately detecting small but critical objects in driving scenes (\eg, \textit{traffic light}).

\begin{figure*}[t]
\begin{center}
\includegraphics[width=17cm]{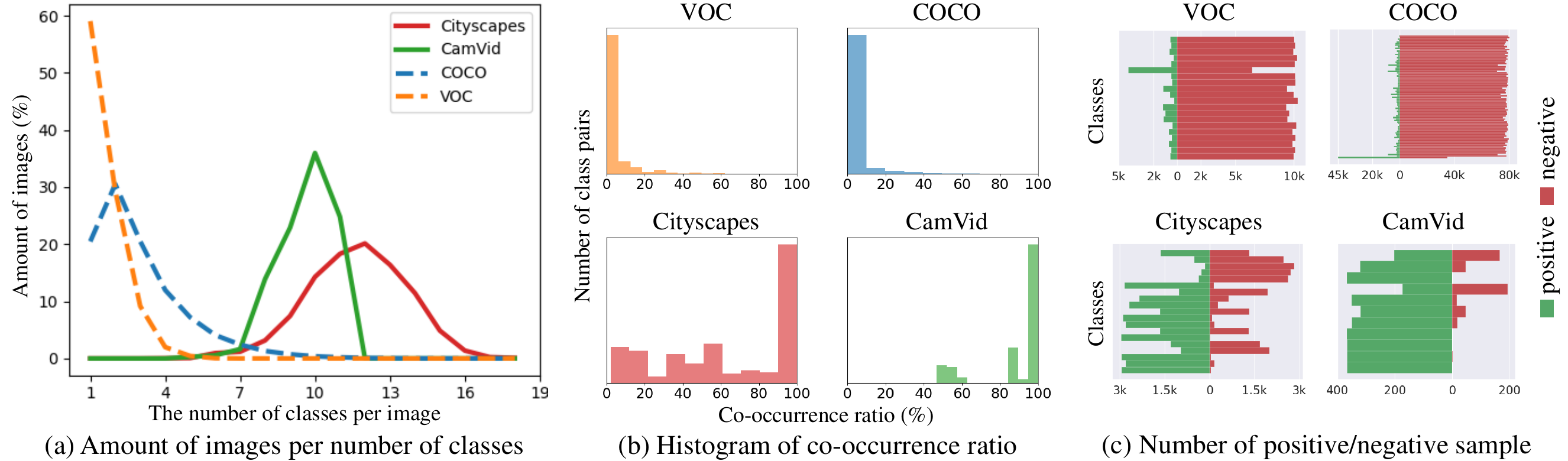}
\end{center}
\caption{Dataset statistics for Cityscapes, CamVid, MS COCO, and PASCAL VOC. (a) Counting the number of images given by the number of classes in a single image. (b) Histogram of co-occurrence ratio between classes. (c) The number of positive and negative images for each class.}
\label{fig:dataset}
\end{figure*}

Next, we propose CARB, which suppresses the erroneous region of pseudo-masks to train the segmentation model. Specifically, we divide the noisy pseudo-mask into consistent and inconsistent regions according to prediction consistency between the segmentation model and CLIP. The inconsistent region contains more false predictions than the consistent region, resulting in a higher loss. This discrepancy in the magnitude of loss values leads to a negative impact on the overall training process. To mitigate this, we propose a strategy to balance the losses from both regions, thereby suppressing the high loss of the inconsistent region.

In summary, we examine the distinct characteristics of driving scenes over the commonly evaluated datasets and highlight the issue of ineffective CAM-based approaches in these scenes. We introduce a new WSSS framework utilizing pseudo-masks generated from CLIP, suggesting global-local view training to handle small-sized objects and CARB to mitigate the negative effects of noisy pseudo-masks. 

We demonstrate that the proposed method achieved 51.8\% mIoU on the Cityscapes dataset, showcasing the potential as a strong WSSS baseline for driving scenes. The effectiveness of the proposed method was confirmed on CamVid representing a small-scale dataset, and on WildDash2 containing more visually challenging scenes (\eg, diverse weather and lighting conditions). Owing to its advantage in performance and simple training, our method can serve as a valuable baseline for future research to address the challenges of WSSS in the driving scenes.

\section{Related Work}
\subsubsection{Earlier Works in WSSS.} Most WSSS techniques using image-level labels utilize CAM~\cite{zhou2016learning}. Due to its sparse coverage, recent studies have focused on expanding discriminative regions~\cite{jiang2019integral, wei2018revisiting, choe2019attention}. In terms of using global-local view, L2G~\cite{jiang2022l2g} strengthened classifier learning by using local attention. This method was also used to widen the discriminative region. Recently, some approaches have attempted to solve co-occurrence problem by incorporating additional information~\cite{lee2021railroad, lee2022weakly, Xie_2022_CVPR}. However, most of existing methods were only evaluated on PASCAL VOC~\cite{everingham2015pascal} or MS COCO~\cite{lin2014microsoft}. \citeauthor{akiva2023single}.~\citeyear{akiva2023single} conducted evaluations on more complex datasets like Cityscapes~\cite{cordts2016cityscapes} and ADE20k~\cite{zhou2019semantic}, but only revealed the performance limitations of existing WSSS studies. \citeauthor{wang2020deep}.~\citeyear{wang2020deep} introduced a clustering-based approach in driving scene datasets, while they only achieved a marginal improvement. Unlike most WSSS studies, we analyze distinct characteristics of the driving scene datasets compared to existing benchmark datasets and suggest a new direction for WSSS in driving scene scenarios.

\subsubsection{CLIP-based Segmentation.} CLIP~\cite{radford2021learning} is a framework trained on a large amount of image-text pairs. Several attempts have been made to utilize the characteristics of the multimodal embedding space in the field of segmentation~\cite{ding2022decoupling,Wang_2022_CVPR}. In WSSS, CLIMS~\cite{Xie_2022_CVPR} employed the embedding spaces by optimizing the mask based on the similarity between masked image and text embedding. CLIP-ES~\cite{Lin_2023_CVPR} generates seed masks in Grad-CAM manner~\cite{selvaraju2017grad}. Then, it refines masks with class-wise attention-based affinity of the CLIP image encoder.

Existing studies~\cite{li2022languagedriven, xu2021} have also shown a significant improvement in zero-shot and few-shot segmentation by leveraging CLIP's zero-shot ability. Recently, MaskCLIP~\cite{zhou2022extract} has been proposed to create dense masks from CLIP with category information at the dataset level rather than the image level. We employ MaskCLIP to extract dense labels from images and further propose a training strategy for handling the noise present in its pseudo-masks.

\begin{figure*}[t!]
\centering
\includegraphics[width=17cm]{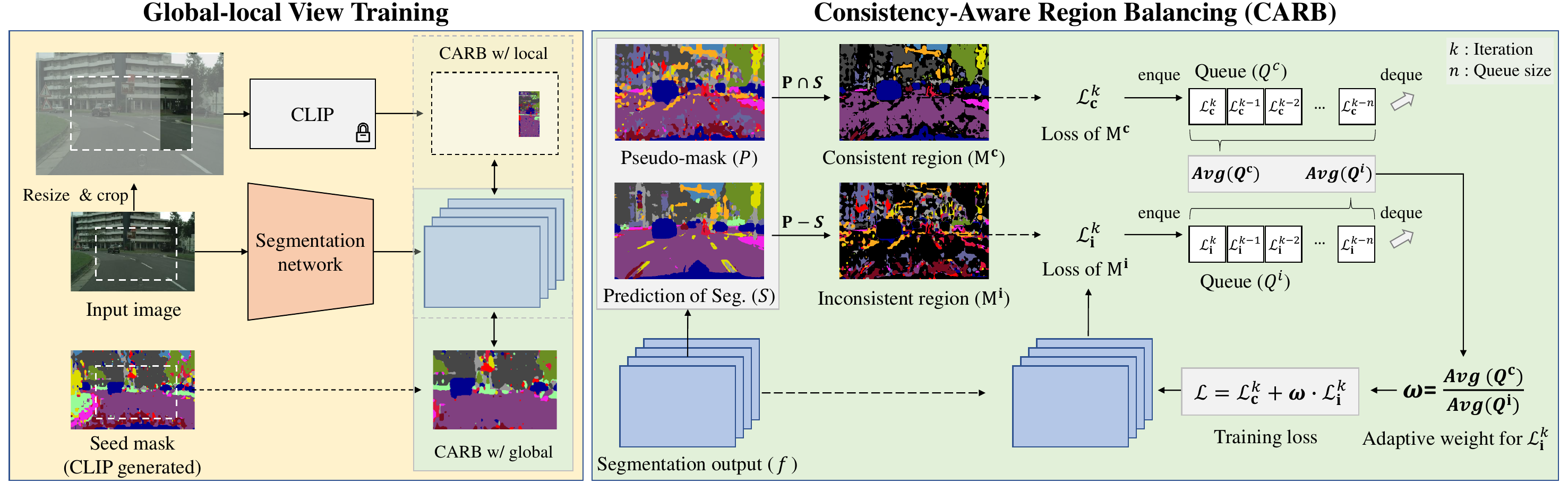}
\caption{Overall framework of proposed method. (Global-local View Training) CLIP gives different pseudo masks for cropping and resizing. (CARB) The pseudo-mask is divided into the consistent / inconsistent regions and the high loss of inconsistent regions is suppressed via adaptive region balancing.}
\label{fig:framework}
\end{figure*}

\subsubsection{Uncertainty Estimation.} Estimating the uncertainty~\cite{DBLP:conf/nips/KendallG17} has been discussed in deep learning since deep neural networks learn to approximate probabilistic models. Focusing on semantic segmentation, the \citeauthor{feng2022dmt} utilizes an ensemble of models that are initialized differently to separate the uncertainty region. In a similar vein, several methods~\cite{oh2021background,DBLP:conf/aaai/ZhangXWSH20} utilized a combination of confidence thresholding and consistency between the CRF-refined mask and the original mask to define a reliable region. ST++~\cite{yang2022st++} identifies reliable images by utilizing the results of previous checkpoints. Recently, several methods suggest pixel-level entropy to measure the pixel-level uncertainty~\cite{NEURIPS2020_f73b76ce,Huynh:CVPR22,li2022uncertainty, wang2022semi}.

\section{Statistics of Datasets}
In this section, to identify the cause of the poor performance of existing WSSS methods on driving scenes, we compare the characteristics of two types of datasets: standard benchmark datasets (\eg, PASCAL VOC and MS COCO) and driving scene datasets (\eg, Cityscapes and CamVid). Specifically, we investigate the histograms of 1) the number of classes per image, 2) the co-occurrence ratio between classes, and 3) the number of positive/negative samples per class (Fig.~\ref{fig:dataset}). 

The distinct difference between the two types of datasets is the number of classes in a single image. Although existing benchmark datasets have only one or two classes in most images, driving scene datasets typically contain eight or more classes in a single image, as in Fig.~\ref{fig:dataset}~(a). Next, we calculate the frequency ratio of co-occurrence between every pair of classes and plot a histogram of those ratios in Fig.~\ref{fig:dataset}~(b). Even worse, these classes in driving scenes often appear together, causing contextual bias. It is clearly different from PASCAL VOC and MS COCO.

Another critical point is the scarcity of negative samples in driving scene datasets. Negative samples are important learning signals for training the image classifier. As shown in Fig.~\ref{fig:dataset}~(c), existing datasets have a sufficient number of negative samples, but some classes in driving scene datasets have extremely few or zero negative samples. Most seriously, \textit{road} and \textit{car} in CamVid always appear in all training images and cannot be distinguished using only image-level labels. Understanding these characteristics is essential for developing productive approaches for WSSS using image-level labels in driving scene applications.

\begin{figure}[t]
\begin{center}
\includegraphics[width=8cm]{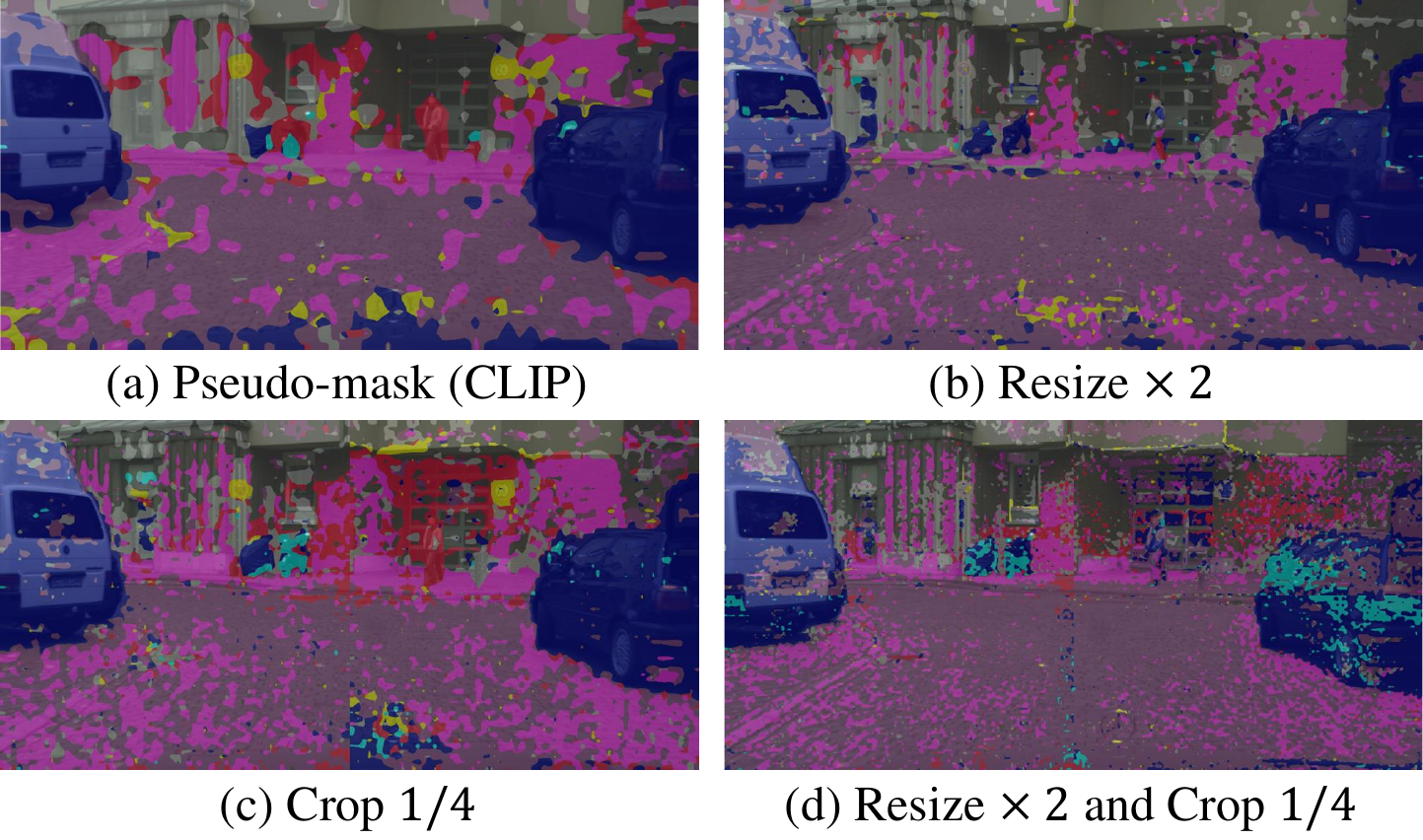}
\end{center}
\caption{Pseudo-masks after resizing and cropping. (a) The original CLIP mask. (b) CLIP mask with resize ratio 2. (c) The concatenation of quarter-size cropped CLIP masks (d) The mask applying both operations. For visual clarity, we modify color palette of motorcycle to \textit{cyan} in this figure.}
\label{fig:mask_resizencrop}
\end{figure}

\begin{figure}[t]
\begin{center}
\includegraphics[width=8cm]{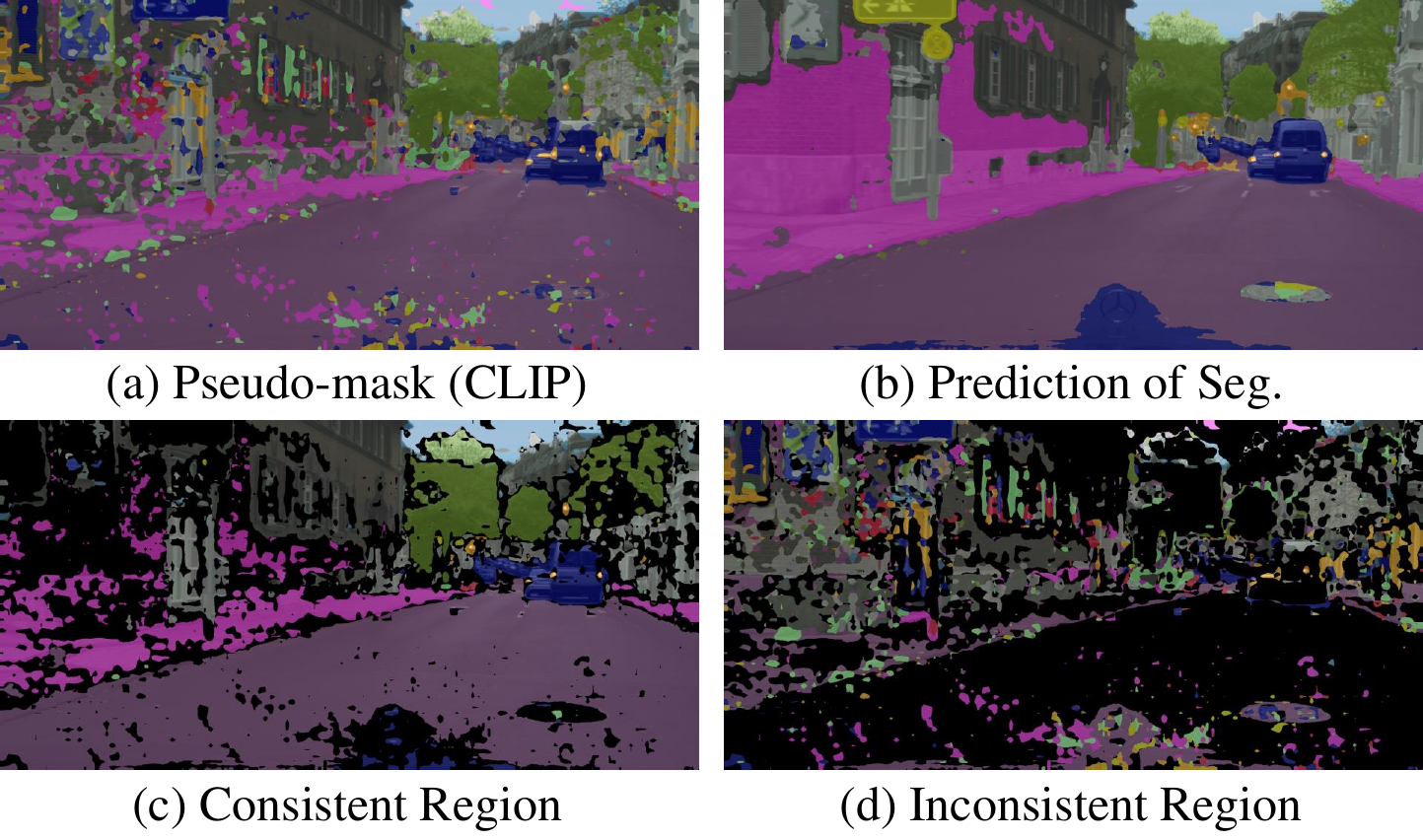}
\end{center}
\caption{The characteristics of two different masks. (a) The mask from CLIP contains small and blob-like noisy regions. (b) The output mask from the segmentation network is more systematic. We identify reliable regions (c) based on prediction consistency between (a) and (b).}
\label{fig:mask_character}
\end{figure}

\begin{figure}[t]
\begin{center}
\includegraphics[width=\linewidth]{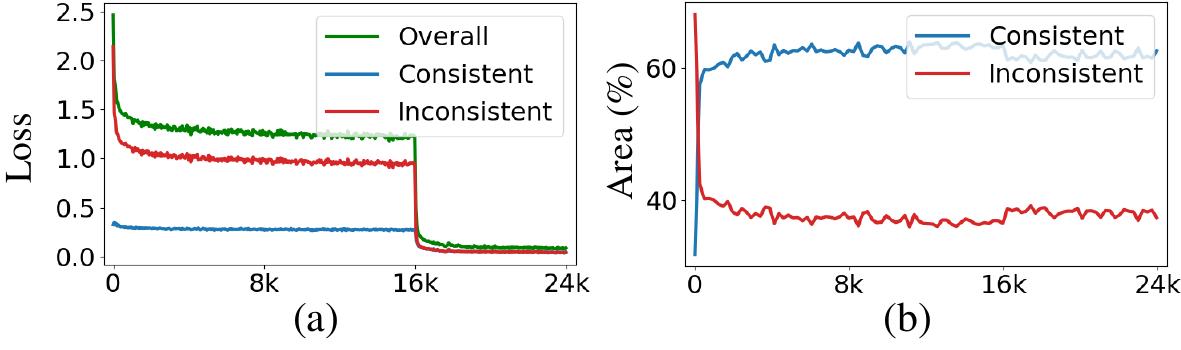}
\end{center}
\caption{Changes in (a) loss and (b) area of consistent/inconsistent regions during training. Adaptive region balancing is applied from 16K iteration, affecting the training dynamics.}
\label{fig:fig_loss1}
\end{figure}

\section{Method}

\subsection{Global-local View Training}
Owing to the specific nature of driving scenes, certain classes such as \textit{roads} are consistently large, and others like \textit{traffic light} remain small in size. Since the driving scenes capture road environments with a wide range of depth in each image, object sizes vary significantly with distance even within the same class, such as \textit{car}. In Fig.~\ref{fig:mask_resizencrop}~(a), we observe that the pseudo-masks generated by the CLIP model exhibit notably high quality for relatively large objects but poor quality for small objects. We conjecture that this performance degradation may occur from the training mechanism of CLIP (\ie, it mainly concentrate on salient objects corresponding to text prompt rather than small objects).

Building upon this observation, we manipulate the relative object size within the input by adjusting the \emph{image scale} and \emph{field-of-view} (FOV). We then analyze the resultant changes in the pseudo-mask obtained from CLIP. In Fig.~\ref{fig:mask_resizencrop}~(b), when the input is scaled to twice its original size, the pseudo-mask exhibit more accurate and fine-grained results along the object boundaries. Additionally, reducing the FOV by half (\ie, the network only observes one-quarter of the input at a time) leads to noticeable changes in the pseudo-mask, particularly pronounced for smaller objects such as \textit{motorcycle} (\cf, Fig~\ref{fig:mask_resizencrop}~(c)). This simple case study unveils distinctive characteristics associated with each adjustment. 

Summarizing, we observed distinct responses of CLIP to cropping (changing FOV) and resizing (changing scales). (1) Resizing improves the localization at fine-grained areas such as edges. (2) Cropping enhances the classification of small objects. Capitalizing on the distinctive effects of cropping and resizing, we synergistically incorporate both augmentations into our approach. By jointly leveraging these functions, we enhance pseudo-mask performance, particularly for small objects. 

Inspired by this observation, we have developed a new method called \textit{Local View Sampling}. This technique leverages the conventional augmented input known as the global view, commonly used for training segmentation networks. We extract a patch of a specific size (typically small) from an arbitrary position inside the global view. Then, the patch is resized randomly before passing it through CLIP. We obtain the local pseudo-mask by calculating the similarity of local features from the image encoder and text embedding as follows:

\begin{equation}
\label{eq_local_mask}
{\small
\mathbf{M}^\mathbf{l} = \arg \max(\frac{\boldsymbol{F}^\mathbf{l} \cdot \boldsymbol{t}}{\left \| \boldsymbol{F}^\mathbf{l} \right \| \cdot \left \| \boldsymbol{t} \right \|}),
}
\end{equation} %
where $\boldsymbol{F}^\mathbf{l}$ is the feature from CLIP with a local view image, $\boldsymbol{t}$ is text embedding of CLIP. The local view only contains semantic information in a small, confined, area, so it can fully exploit locality from CLIP. This empowers the pseudo-mask of the local view to better focus on small objects. We leverage the masks of both views to train the segmentation model. The loss for the global-local view training is computed by cross-entropy loss for each region as follows:
\begin{equation}
\label{eq_loss_loc}
{\small
\mathcal{L}_\mathbf{l} = - \frac{1}{|\mathbf{M}^\mathbf{l}|} \underset{i,j \in local}{\sum} \text{ } y^\mathbf{l} \log {f_{i,j}},
}
\end{equation}

\begin{equation}
\label{eq_loss_glo}
{\small
\mathcal{L}_\mathbf{g} = - \frac{1}{|\mathbf{M}^\mathbf{g}|} \underset{i,j}{\sum} \text{ } y^\mathbf{g} \log {f_{i,j}},
}
\end{equation}%
where $y^\mathbf{l}$ and $y^\mathbf{g}$ are one-hot labels of  $\mathbf{M}^\mathbf{l}$ and $\mathbf{M}^\mathbf{g}$, respectively. $\mathbf{M}^\mathbf{g}$ and $\mathbf{M}^\mathbf{l}$ are the global pseudo-mask and the local pseudo-mask, respectively. $f \in \mathbb{R}^{C \times H \times W}$ is the probability of segmentation network. The total loss for the global-local view training is $\mathcal{L} = \mathcal{L}_\mathbf{g} + \mathcal{L}_\mathbf{l}$.

\subsection{Consistency-aware Region Balancing} \label{sec:separation} 
We identify noisy regions in the pseudo-mask created by CLIP. Fig.~\ref{fig:mask_character} showcases an example of a pseudo-mask containing small and blob-like noisy regions, randomly scattered on the image.

Conversely, training a segmentation network with pseudo-masks removes the randomly scattered noise of CLIP's pseudo-mask in the output, resulting in systematic predictions. (\eg, \textit{road} in Fig.~\ref{fig:mask_character} (b)) However, the segmentation prediction has misclassified pixels that were originally correct in the pseudo-mask. In particular, we observe that the trained segmentation model produces an indistinct boundary of the object even worse than the pseudo-mask generated by CLIP (\eg, \textit{sidewalk} in Fig.~\ref{fig:mask_character} (b)).

Owing to the unique properties of the trained segmentation model and CLIP, we leverage both models to benefit from their respective advantages. However, since the prediction of the segmentation model is already incorporated in the model, directly computing the loss using the segmentation prediction does not provide new evidence for training. To address this, we indirectly employ the segmentation prediction to distinguish the pixels of the pseudo-mask from CLIP. Specifically, utilizing prediction consistency, we regard the pixel as reliable if they are consistent and noisy if the predictions from the two models are inconsistent:

\begin{equation}
\label{eq_region_con}
{\small
\mathbf{M}^\mathbf{c} = \{P_{i,j} | P_{i,j} = S_{i,j}\},
}
\end{equation}
\begin{equation}
\label{eq_region_inc}
{\small
\mathbf{M}^\mathbf{i} = \{P_{i,j} | P_{i,j} \neq S_{i,j}\},
}
\end{equation}%
where $\mathbf{M}^\mathbf{c}$ and $\mathbf{M}^\mathbf{i}$ correspond to consistent and inconsistent regions, respectively. $P \in C^{H \times W}$ and $S \in C^{H \times W}$ are the pseudo-mask from CLIP and the prediction of the segmentation network, where $C$ is a set of classes. Furthermore, we apply label filtering when generating $S$ to prevent misprediction with non-existent classes in the image. 

The consistent and inconsistent regions are recalculated in each iteration to update the segmentation model. As training progresses, we notice that the size of the consistent region changes, resulting in performance improvement of the segmentation model (Fig.~\ref{fig:fig_loss1}~(b)). 

To understand the effects of consistent and inconsistent regions, we separately calculate the cross-entropy loss of each side: 
\begin{equation}
\label{eq_loss_con}
{\small
\mathcal{L}_\mathbf{c} = - \frac{1}{|\mathbf{M}^\mathbf{c}|} \underset{i,j}{\sum} \text{ } y^\mathbf{c} \log {f_{i,j}},
}
\end{equation}

\begin{equation}
\label{eq_loss_inc}
{\small
\mathcal{L}_\mathbf{i} = - \frac{1}{|\mathbf{M}^\mathbf{i}|} \underset{i,j}{\sum} \text{ } y^\mathbf{i} \log {f_{i,j}},
}
\end{equation}%
where $y^\mathbf{c}$ and $y^\mathbf{i}$ are one-hot labels of  $\mathbf{M}^\mathbf{c}$ and $\mathbf{M}^\mathbf{i}$, respectively. $f \in \mathbb{R}^{C \times H \times W}$ is the probability of segmentation network. We observe that inconsistent regions have much higher loss values than consistent regions in Fig.~\ref{fig:fig_loss1} (a). If we treat the training loss equally across all regions, the network is overly influenced by the high loss produced from the inconsistent regions. Therefore, we suggest assigning different weights to the losses of consistent and inconsistent regions while taking into account the noise level of the data. It helps prevent the conventional cross-entropy loss from being vulnerable to noise in the training data.

To this end, we devise an adaptive region balancing method that dynamically adjusts the loss of the inconsistent region by monitoring the loss profiles in both the consistent and inconsistent regions during training. Specifically, we introduce two fixed-size queues which track the losses of the two regions, denoted as $\mathcal{L}_\mathbf{c}$ and $\mathcal{L}_\mathbf{i}$, respectively. We then compute the average loss from each queue. We use the ratio of two average losses as the weight for the cross-entropy loss of the inconsistent region, denoted as $w$, which is multiplied by the loss of the inconsistent region. The CARB training loss is $\mathcal{L} = \mathcal{L}_\mathbf{c} + w \cdot \mathcal{L}_\mathbf{i}$. This balancing ensures that the training is less influenced by the inconsistent region. 

While one might consider that the loss from inconsistent regions can be simply neglected, our observations reveal that the inconsistent regions still possess useful learning signals. Notably, we observe that labels of highly correlated object classes (\ie, the classes sharing visual properties like \textit{bus} and \textit{car}) exist within the inconsistent region. Overlooking those pixels impedes the label imbalance problem. The Cityscapes dataset, for instance, includes classes like \textit{rider} (a subset of \textit{person}), \textit{bus}, and \textit{truck} (subsets of \textit{car}) that are susceptible to such confusion. Considering these challenges, we present a region-balancing method designed to harness meaningful information even from inconsistent regions.

\subsubsection{Overall Training.}
The proposed method consists of two stages. In the first stage, we warm up the baseline segmentation model with global and local views generated from CLIP masks. This step ensures that the segmentation network sufficiently learns the regular patterns of the target dataset. In the second stage, we refine the segmentation network utilizing CARB. We apply CARB for both global and local views.

\section{Experiments}
\subsection{Experimental Setup}
\subsubsection{Dataset \& Evaluation Metric.} For performance evaluation, we utilized the well-known Cityscapes~\cite{cordts2016cityscapes}, CamVid~\cite{brostow2009semantic}, and WildDash2~\cite{Zendel_2022_CVPR}, which are autonomous driving datasets. The Cityscapes dataset consists of 2,975 training, 500 validation, and 1,525 test images with fine annotation. It contains a total of 30 classes, and 19 classes are evaluated for public assessment while the rest are void. The CamVid dataset consists of 367 training, 101 validation, and 233 test images, containing a total of 32 classes. In our experiments, only 11 classes are evaluated by following the convention of previous research~\cite{wang2020deep}. The WildDash2 dataset consist of 3,618 training, 638 validation, and 812 test images, containing a total of 25 classes. In all experiments, we solely utilized image-level labels for training. The image-level labels are acquired from pixel-level labels of each dataset. Mean Intersection over Union (mIoU) was used as the evaluation criterion, a popular and standard metric for semantic segmentation.

\subsubsection{Implementation Detail.} 
We employed ViT-B/16~\cite{dosovitskiyimage} as the image encoder for the CLIP, and ResNet50~\cite{he2016deep}-based DeepLab-ASPP~\cite{chen2017deeplab} as the segmentation network. The last convolutional layer of ASPP was replaced with text embedding of CLIP. The segmentation network was initialized with an ImageNet pre-trained model provided by MMSeg~\cite{mmseg2020}. Considering class definition and object words connoting actual objects, we replaced the \textit{vegetation} and \textit{terrain} class names with \textit{tree} and \textit{grass}, respectively. Furthermore, we changed the \textit{person} class to \textit{pedestrian}, since it is a superset of the \textit{rider}. For generating pseudo-masks from CLIP, we adopt MaskCLIP~\cite{zhou2022extract}. 

\begin{table}[]
\normalsize
\centering
{\small
\begin{tabular}{@{}lc@{}}
\toprule
Method          & \multicolumn{1}{c}{mIoU}  \\ \midrule
Base                                & \multicolumn{1}{c}{40.1} \\
\multicolumn{1}{l}{+ CARB}           & \multicolumn{1}{c}{45.7} \\
\multicolumn{1}{l}{+ Local}          & \multicolumn{1}{c}{45.1} \\ 
\multicolumn{1}{l}{+ Local + CARB}   & \multicolumn{1}{c}{50.6} \\ 
\multicolumn{1}{l}{+ Dual}           & \multicolumn{1}{c}{45.8} \\ 
\multicolumn{1}{l}{+ Dual + CARB}    & \multicolumn{1}{c}{\textbf{52.1}} \\

\bottomrule
\end{tabular}
}
\caption{Ablation study of the proposed modules. The accuracy (mIoU) is evaluated on the Cityscapes validation set. The best score is in \textbf{bold} throughout all experiments.} 
\label{tab:split}
\end{table}
 
\subsection{Ablation Study}
\label{sec:ablation}
\subsubsection{Effects of Each Module.} We evaluate the effectiveness of each component of our method in Tab.~\ref{tab:split}. When we train the segmentation model with additional local view sampling ($Local$), it shows a remarkable improvement of 5.0\%p. This implies that additional information from local patches through cropping and resizing provides rich learning signals. CARB alone contributed an impressive improvement of 5.6\%p, indicating that adaptively re-weighting the loss according to its reliability plays a critical role in learning with noisy pseudo-masks. By combining local view sampling and CARB ($Local + CARB$), a substantial improvement of 10.5\%p was achieved. Instead of $Local$, we added a slight modification named $Dual$, by re-creating the mask of global views depending on the augmentation using CLIP for each iteration. This modification yielded a 0.7\%p gain over $Local$. Interestingly, our $Dual+CARB$ method shows a 1.5\%p improvement compared to $Local + CARB$, indicating synergy between various-sized mask creations and our noise-handling strategy. 

\begin{figure}[t]
\begin{center}
\includegraphics[width=\linewidth]{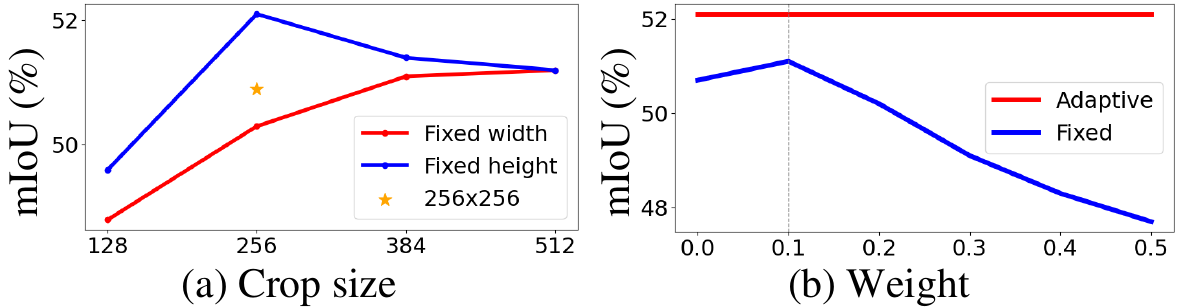}
\end{center}
\caption{Segmentation results (mIoU) on Cityscapes validation set depending on (a) the crop size (b) the weight. We set the length of one side to 512 and varied the length of the other side between 128 and 512. Yellow star indicates the experiment using $256 \times 256$ patch.}
\label{fig:ablation}
\end{figure}

\begin{table}[]
\normalsize
\centering
{\small
\begin{tabular}{@{}lrr@{}}
\toprule
\multicolumn{1}{c}{Method}                              & \multicolumn{1}{c}{val} & \multicolumn{1}{c}{test} \\ \midrule
\multicolumn{1}{l}{DeepLab-ASPP (Full supervision)} & 78.3 & \multicolumn{1}{r}{75.8} \\ \midrule
\multicolumn{1}{l}{AffinityNet} &  8.2 & \multicolumn{1}{r}{-} \\
\multicolumn{1}{l}{SEAM} & 17.3 & \multicolumn{1}{r}{-} \\
\multicolumn{1}{l}{1-Stage} & 11.8 & \multicolumn{1}{r}{-} \\
\multicolumn{1}{l}{\citeauthor{wang2020deep}} & 24.2 & \multicolumn{1}{r}{24.9}          \\ 
\multicolumn{1}{l}{CAM} & 33.0 & \multicolumn{1}{r}{32.2} \\ 
\multicolumn{1}{l}{AMN} & 17.5 & \multicolumn{1}{r}{17.8} \\ 
\multicolumn{1}{l}{CLIMS } & 18.1 & \multicolumn{1}{r}{18.0} \\
\multicolumn{1}{l}{CLIP-ES } & 35.4 & \multicolumn{1}{r}{35.0} \\ \midrule
\multicolumn{1}{l}{Ours}                                & \textbf{52.1}     & \multicolumn{1}{r}{\textbf{51.8}} \\ \bottomrule

\end{tabular}
}
\caption{Segmentation results (mIoU) on Cityscapes.}
\label{tab:seg_cityscapes}
\end{table}

\subsubsection{Effects of the Crop Size and Resize Ratio.} In our empirical investigation, it was consistently observed that vertically long rectangular patches exhibited superior performance in terms of crop sizes compared to patches of other sizes. This finding is supported by Fig.~\ref{fig:ablation}. We conjecture this tendency in the driving scene dataset comes from vertically long structures such as \textit{pole} and \textit{traffic light}. Also, the 512$\times$512 patches for local views are more effective than the 256$\times$256. These experiments suggest that, while the local view represents a smaller portion of the overall scene, excessively small sizes may not benefit from the attention layers equipped in CLIP.

We evaluated our local view sampling under variable resize ratios. Under the fixed ratio from 0.5 to 2.0, we observe the best performance of 52.1\% at the ratio of 1.0 while a significant drop with other ratios. However, when focusing on classwise performances under various resize ratios, we confirmed that a large resize rate benefits the performance of small classes, such as \textit{traffic light} and \textit{rider}. Meanwhile, the performance of the large classes, such as \textit{sidewalk} and \textit{wall}, is decreased. Due to the performance trade-off across different classes, we set a random value between 1.0 and 2.0 as the resize ratio. Our choice leads to the overall best performance in both small and large classes.

\subsubsection{Effects of Adaptive Region Balancing.} We compare our adaptive region balancing strategy with a fixed weighting strategy, where the loss weight for the inconsistent region ($w$) is set to a specific value (\cf, Fig.~\ref{fig:ablation} (b)). When changing the fixed weight $w$ gradually from 0 to 0.5, we observe the best score at the weight of 0.1 and a significant drop with other values. Although the highest performance of the fixed weight strategy is similar to that of our method (51.06\% for fixed strategy and 52.1\% for our method), it requires a hyperparameter search for the optimal weight using the validation dataset. In contrast, our method does not require such a search, making it more suitable for WSSS scenario.

\subsection{Quantitative Comparisons}
\subsubsection{Remarks on Comparisons.} Existing WSSS methods focus on handling object-centric datasets such as PASCAL VOC 2012. Therefore, their methods are primarily designed to distinguish relatively simple object shapes with similar scales, which is still a valuable research direction. Given this dataset mismatch, direct comparisons between our method and existing WSSS approaches might not be entirely fair, as they cater to distinct dataset characteristics. Nevertheless, by adapting established WSSS methods to driving scenes, we intend to show that the existing framework is ineffective for our application scenario.

\begin{figure*}[t!]
\centering
\includegraphics[width=16cm]{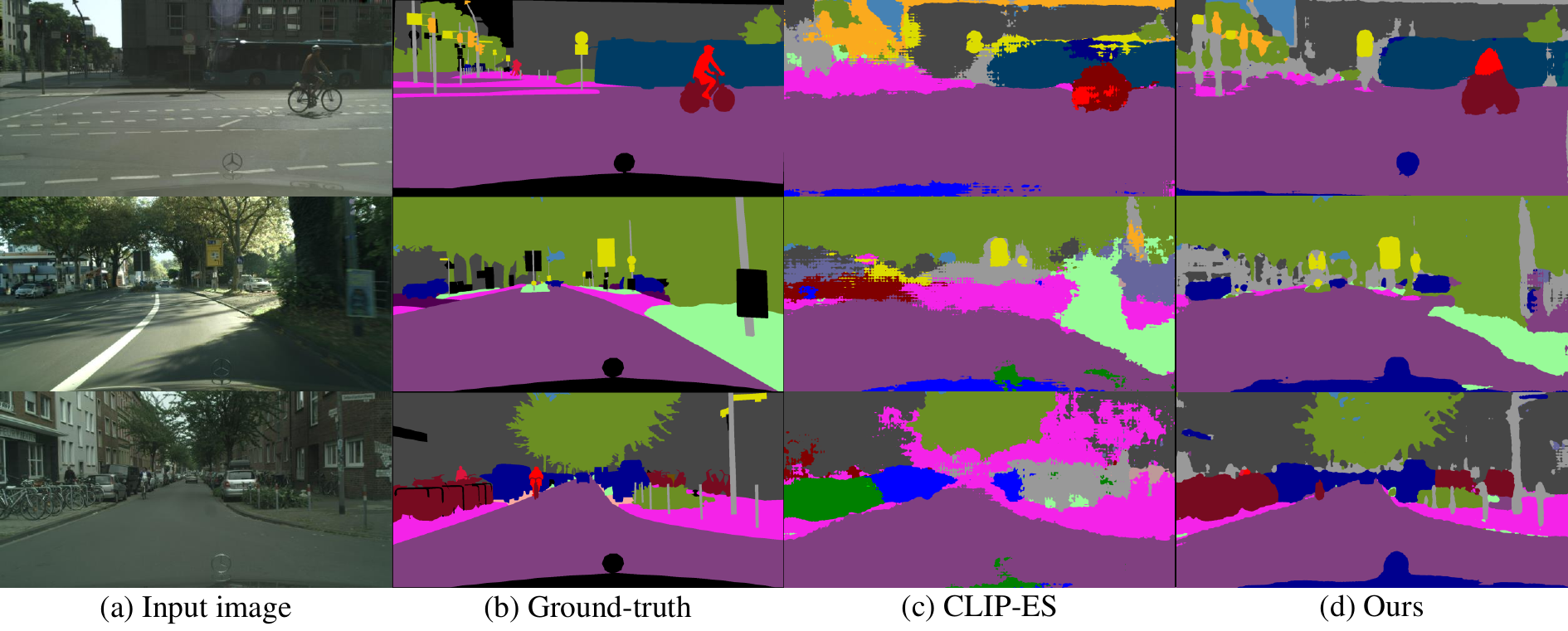}
\caption{Qualitative results on Cityscapes validation set. (a) Input image, (b) Ground-truth, (c) CLIP-ES, and (d) Our method.}
\label{fig:qualitative_city}
\end{figure*}

\subsubsection{Existing WSSS Methods.} Existing methods can be partitioned into CAM-based methods (\ie, an image classifier for generating the pseudo-masks) and CLIP-based methods (\ie, CLIP for generating the pseudo-masks). Among them, we choose several representative methods such as (1) AffinityNet~\cite{ahn2018learning}, (2) SEAM~\cite{wang2020self}, (3) 1-Stage~\cite{araslanov2020single}, (4) SEC~\cite{kolesnikov2016seed}, (5) Wang et al.~\cite{wang2020deep}, (6) CAM~\cite{zhou2016learning}, and  (7) AMN~\cite{lee2022threshold}. To compare with CLIP-based WSSS, we reproduce the driving scene results of (8) CLIMS~\cite{Xie_2022_CVPR} and (9) CLIP-ES~\cite{Lin_2023_CVPR}, both of which use the same level of information, CLIP, as our method. 

\begin{table}[]
\normalsize
\centering
{\small
\begin{tabular}{@{}lrr@{}}
\toprule
\multicolumn{1}{c}{Method}          & \multicolumn{1}{c}{val}   & \multicolumn{1}{c}{test} \\ \midrule
\multicolumn{1}{l}{DeepLab-ASPP (Full supervision)}     & 81.6                      & \multicolumn{1}{r}{74.9}              \\ \midrule
\multicolumn{1}{l}{SEC} & - &  \multicolumn{1}{r}{2.5}              \\
\multicolumn{1}{l}{AffinityNet} & 11.0 & \multicolumn{1}{r}{15.5} \\
\multicolumn{1}{l}{\citeauthor{wang2020deep}} & 23.5 & \multicolumn{1}{r}{30.4} \\
\multicolumn{1}{l}{CAM} & 9.6 & \multicolumn{1}{r}{6.6} \\
\multicolumn{1}{l}{AMN} & 10.7 & \multicolumn{1}{r}{7.6} \\ 
\multicolumn{1}{l}{CLIMS} & 2.7 & \multicolumn{1}{r}{4.3} \\ 
\multicolumn{1}{l}{CLIP-ES} & 41.7 & \multicolumn{1}{r}{39.6} \\ \midrule
\multicolumn{1}{l}{Ours} & \textbf{55.7} & \multicolumn{1}{r}{\textbf{50.5}} \\ \bottomrule
\end{tabular}
}

\caption{Segmentation results (mIoU) on CamVid.}
\label{tab:seg_camvid}
\end{table}

\subsubsection{Cityscapes.} Tab.~\ref{tab:seg_cityscapes} presents the performance of our proposed CARB compared to other methods in driving scenes. Specifically, our approach achieves 51.8\% on the Cityscapes test set, which outperforms the \citeauthor{wang2020deep} by 26.9\%p and
previous CLIP-based WSSS technique by 16.8\%p. Additionally, we observed that our method consistently performs better than CLIP-ES in every class. Fig.~\ref{fig:qualitative_city} showcases qualitative examples of segmentation results on the Cityscapes. Notably, CARB successfully eliminates misclassified \textit{sidewalk} regions on the \textit{sky} class (see the first and second rows). These results visually confirm that our method correctly captures each class and successfully reduces the prediction errors. 

\subsubsection{CamVid.} The CamVid dataset has a much smaller number of training images compared to Cityscapes, with only 367 images. Additionally, it is not possible to differentiate between the \textit{car} and \textit{road} classes using only image-level labels, as they appear in all images. However, our method can distinguish them by utilizing the pre-trained image-text information from the CLIP model. Tab.~\ref{tab:seg_camvid} shows that the performance of CAM-based methods (\eg, SEC, AffinityNet, \citeauthor{wang2020deep}, and CLIMS) is considerably low while our method achieves significantly higher performance. This demonstrates that our proposed approach can address the problem even when the scale of the dataset is small and has severe contextual bias. 

\subsubsection{WildDash2.} Since the WildDash2 dataset possesses extremely high diversity, it is generally challenging even for the fully supervised model. A classifier-based WSSS method such as CLIMS performs poorly, 1\% in mIoU, which is worse than a random guess. This poor performance is caused by difficulties in training the classifier due to class imbalance and complex class distribution. Since CLIP-ES and our method are built upon CLIP for generating the pseudo masks, both methods provide relatively reasonable performances. Our method achieves considerably high performance compared to CLIP-ES, with the performance gain primarily observed in small classes such as \textit{billboard}, \textit{rider}, \textit{bicycle}, and \textit{road marking}.

\begin{table}[]
\centering
{\small
\begin{tabular}{@{}lr@{}}
\toprule
\multicolumn{1}{c}{Method}                              & \multicolumn{1}{c}{Result} \\ \midrule
\multicolumn{1}{l}{DeepLab-ASPP (Full supervision)}     & \multicolumn{1}{r}{54.0}          \\ \midrule
\multicolumn{1}{l}{CAM} & \multicolumn{1}{r}{15.2} \\
\multicolumn{1}{l}{AMN} & \multicolumn{1}{r}{18.8} \\
\multicolumn{1}{l}{CLIMS} & \multicolumn{1}{r}{1.0} \\
\multicolumn{1}{l}{CLIP-ES} & \multicolumn{1}{r}{24.7}          \\ \midrule
\multicolumn{1}{l}{Ours}                                & \multicolumn{1}{r}{\textbf{32.2}} \\ \bottomrule
\end{tabular}
}
\caption{Segmentation results (mIoU) on WildDash2 \textit{val} set.}
\label{tab:seg_coarse}
\end{table}

\section{Conclusion}
This paper addressed the limitations of conventional CAM-based, weakly-supervised semantic segmentation (WSSS) methods when handling the driving scene datasets. To break the performance bottleneck of the CAM-based methods, we utilized CLIP as the pseudo-mask generator. Then, we proposed global-local view training, which exploits the characteristics of CLIP generating diverse masks depending on the relative object sizes. We also propose a novel training strategy, namely consistency-aware region balancing (CARB). It distinguishes between reliable and noisy regions utilizing prediction consistency and then suppresses the latter regions during training. By incorporating  these two components, our method successfully (1) learns to segment small objects and (2) heavily relies on reliable regions while effectively handling challenging objects from noisy regions. Extensive experiments demonstrate that each component of our method contributes to achieving new state-of-the-art performances on the Cityscapes, CamVid, and WildDash2 datasets in WSSS. Our study introduces a new approach addressing the challenges posed by driving datasets and suggests a promising direction for future research in WSSS.

\section*{Acknowledgments}
This research was supported by the Basic Science Research Program through the National Research Foundation of Korea (NRF) funded by the MSIP (NRF-2022R1A2C3011154, RS-2023-00219019, RS-2023-00240135) and MOE (NRF-2022R1A6A3A13073319), the IITP grant funded by the Korea government(MSIT) (No. 2019-0-00075, Artificial Intelligence Graduate School Program(KAIST)), KEIT grant funded by the Korea government(MOTIE) (No. 2022-0-00680, 2022-0-01045, 2021-0-02068, Artificial Intelligence Innovation Hub) and Samsung Electronics Co., Ltd (IO230508-06190-01).

\end{document}